\documentclass[runningheads,fleqn,a4]{llncs}

\usepackage{latexsym}
\usepackage{amssymb}
\usepackage{amsmath}
\usepackage[all]{xy}
\usepackage{hyperref}
\usepackage{wrapfig}

\usepackage{listings}
\newcommand{\ignore}[1]{} 

\makeatletter
\newenvironment{prog}{\vspace{0.7ex}\par
\setlength{\parindent}{0.7cm}
\obeylines\@vobeyspaces\tt}{\vspace{0.7ex}\noindent
}
\makeatother
\newcommand{\startprog}{\begin{prog}}
\newcommand{\stopprog}{\end{prog}\noindent}

\newcommand{\id}{{\mathit{id}}} 

\newcommand{\sleq}{\leqslant}

\def\defemb#1#2{\expandafter\def\csname #1\endcsname
                              {\relax\ifmmode #2\else\hbox{$#2$}\fi}}
\defemb{cA}{{\cal A}}
\defemb{cB}{{\cal B}}
\defemb{cC}{{\cal C}}
\defemb{cD}{{\cal D}}
\defemb{cE}{{\cal E}}
\defemb{cF}{{\cal F}}
\defemb{cG}{{\cal G}}
\defemb{cH}{{\cal H}}
\defemb{cI}{{\cal I}}
\defemb{cJ}{{\cal J}}
\defemb{cL}{{\cal L}}
\defemb{cM}{{\cal M}}
\defemb{cN}{{\cal N}}
\defemb{cO}{{\cal O}}
\defemb{cP}{{\cal P}}
\defemb{cQ}{{\cal Q}}
\defemb{cR}{{\cal R}}
\defemb{cS}{{\cal S}}
\defemb{cT}{{\cal T}}
\defemb{cU}{{\cal U}}
\defemb{cV}{{\cal V}}
\defemb{cX}{{\cal X}}
\defemb{cZ}{{\cal Z}}
\defemb{cW}{{\cal W}}

\newcommand{\var}{\mathsf{var}}

\newcommand{\dom}{{\cD}om}

\newcommand{\cons}{\!:\!}

\newcommand{\ol}[1]{\overline{#1}}  

\def \tuple#1{\langle #1 \rangle}

\long\def\comment#1{}

\newcommand{\mgu}{\mathsf{mgu}}

\newcommand{\subs}{\mathsf{subs}}
\newcommand{\none}{\mathsf{none}}

\newcommand{\sldp}{SL\red{P}DNF}
\newcommand{\expl}{\mathsf{expl}}
\newcommand{\duals}{\mathsf{duals}}
\newcommand{\hits}{\mathsf{hits}}
\newcommand{\mins}{\mathsf{mins}}
\newcommand{\dnf}{\mathsf{dnf}}

\newcommand{\trp}{\mathsf{trp}}
\newcommand{\trc}{\mathsf{trc}}
\newcommand{\q}{\mathsf{chq}}

\usepackage{xcolor}

\newcommand{\red}[1]{{\color{red} #1}}

\newcommand{\blue}[1]{{\color{blue} #1}}

\newcommand{\ch}{\mathsf{C}}
\newcommand{\bigch}{\mathbb{C}}
\newcommand{\ffp}{\mathit{ffp2}}

\begin{document}

\title{Explaining Explanations in\\ 
Probabilistic Logic Programming%
\thanks{This work has been partially supported by grant PID2019-104735RB-C41
funded by MICIU/AEI/ 10.13039/501100011033, by the 
\emph{Generalitat Valenciana} under grant CIPROM/2022/6 
(FassLow), and by TAILOR, a project funded by EU Horizon 2020 
research and innovation programme under GA No 952215.
}
}

\author{Germ\'an Vidal}

\institute{
VRAIN, Universitat Polit\`ecnica de Val\`encia, Spain\\
  \email{gvidal@dsic.upv.es}
}

\titlerunning{Explaining Explanations in\\ 
Probabilistic Logic Programming}

\maketitle

\pagestyle{headings}

\begin{abstract}
The emergence of tools based on artificial intelligence 
has also led to the need of producing explanations 
which are
understandable by a human being. In most approaches, 
the system is considered a \emph{black box}, making 
it difficult to generate appropriate explanations. 
In this work, though, we consider a setting where models
are \emph{transparent}: probabilistic logic 
programming (PLP), a paradigm that combines logic programming for 
knowledge representation and probability to model
uncertainty.
However, given a query, the usual notion of 
\emph{explanation} is associated with a set 
of choices, one for each random variable of the model. 
Unfortunately, such a set does not explain \emph{why}
the query is true and, in fact, it may contain choices 
that are actually irrelevant for the considered query. 
To improve this situation, 
we present in this paper an approach to explaining
explanations which is based on defining a new
query-driven inference mechanism for PLP 
where proofs are labeled with 
\emph{choice expressions}, a compact and
easy to manipulate representation for sets of choices.
The combination of proof trees and
choice expressions allows us to produce comprehensible
query justifications with a causal structure.\\
\begin{quote}
\emph{(*) This preprint has not undergone peer review
      or any post-submission improvements or corrections. 
      The Version of Record of this contribution is published 
      in Programming Languages and Systems (Proceedings of
      APLAS 2024), Springer LNCS, 2024, and is available online 
      at\linebreak \emph{\url{https://doi.org/10.1007/978-981-97-8943-6_7}}}
\end{quote}
\end{abstract}

\section{Introduction} \label{sec:intro}

\emph{Explainable AI} (XAI) \cite{ARSBTBGGM20} is an active 
area of research that includes many different approaches. 
Explainability is especially important in the context of 
decision support systems, where the user often demands to 
know the reasons for a decision. 
Furthermore, the last regulation on
data protection in the European Union \cite{GDPR16} has
introduced a ``right to explanation" for algorithmic decisions.

The last decades have witnessed the emergence of a good number 
of proposals to combine logic programming and probability,
e.g., Logic Programs with Annotated Disjunctions (LPADs)
\cite{VVB04}, CP-logic \cite{VDB09}, ProbLog \cite{dRKT07}, 
Probabilistic Horn Abduction \cite{Poo93},
Independent Choice Logic \cite{Poo97},
PRISM \cite{SK97}, Stochastic Logic Programs
\cite{Mug96}, 
and Bayesian Logic Programs \cite{KR01},
to name a few
(see, e.g., the survey \cite{RS18} and references therein).
Most of these approaches are based on the so-called
\emph{distribution} semantics introduced by Sato \cite{Sat95}.
In this work, we mainly follow the LPAD \cite{VVB04} approach
to probabilistic logic programming (PLP), which has an
expressive power similar to, for example, Bayesian networks \cite{Rig18}.
E.g., the following probabilistic clause:\\[1ex]
\mbox{}\hspace{4ex}$
\blue{heads(X)\cons \red{0.5};\:tails(X)\cons \red{0.5} \leftarrow toss(X),
\neg biased(X).}
$\\[1ex]
specifies that every time a coin $X$ which is not biased
is tossed, it lands on heads with probability 0.5 and on tails
with probability 0.5. Note that only one choice can
be true for a given $X$ (thus ``$;$'' should not be interpreted
as logical disjunction). 
One can say that each instance of the clause above
represents a \emph{random variable} with as many values
as head disjuncts (two, in this case).

Given a program, a \emph{selection} is basically a choice of
values for \emph{all} the random variables represented
in a probabilistic logic program.
Every selection induces a possible
\emph{world}, a normal logic program which is
obtained 
by choosing the head determined by
the selection in each grounding of each probabilistic
clause (and removing its probability). 
For example, given an instance of the clause above for
$X=coin1$, a selection that chooses $heads(coin1)$ will
include the normal clause\\[1ex]
\mbox{}\hspace{4ex}$
\blue{heads(coin1) \leftarrow toss(coin1),
\neg biased(coin1).}
$\\[1ex]
%
In this context, an \emph{explanation} often refers 
to a selection or, equivalently, to the world induced from it.
For instance, the MPE task \cite{SRVKMJ14},
which stands for \textit{Most Probable Explanation}, basically
consists in finding the world with the highest probability given a query (typically denoting a set of observed \emph{evidences}).

A world can be seen indeed as an interpretable model 
in which a given query holds. However, this kind
of explanations also presents several drawbacks.
First, a world might include clauses which are
 irrelevant for the query, thus adding noise to the explanation.
Second, the chain of inferences that proved the query 
 is far from obvious from the explanation (i.e., 
 from the given world). In fact, there can be several 
 different chains of inferences that {explain} why the query 
 is true, each with an associated probability.
Finally, a world (a set of clauses) might be too 
 technical an explanation for non-experts.

Alternatively, some work considers that an explanation 
for a query is given by a set of choices: those that 
are \emph{necessary} to prove the query (see, e.g., \cite{Rig09}). 
Although in this case there is no irrelevant information, 
it still does not have a causal structure. 
Furthermore, a set of choices does not provide an 
intuitive explanation about why the query holds.
%
%
In order to \emph{explain} explanations, we propose in this work
a combination of proof trees---that show the chain of
inferences used to prove a query---and a new representation
for choices that gives rise to a more compact notation.
For this purpose, we make the following contributions:
%
\begin{itemize}
\item First, we introduce an algebra of 
\emph{choice expressions}, a new notation 
for representing sets of choices that 
can be easily manipulated using standard
rules like distributivity, double negation
elimination, De Morgan's laws, etc.

\item Then, we present \sldp-resolution, a query-driven
top-down inference mechanism that extends SLDNF-resolution 
to deal with LPADs. We prove its soundness and completeness
regarding the computation of explanations.

\item Finally, we show how the proofs of 
\sldp-resolution can be used to produce comprehensible 
representations of the explanations of a query, 
which might help the user to understand why this 
query is indeed true.
\end{itemize}

\section{Some Concepts of Logic Programming and PLP} \label{sec:prelim}

In this section, we introduce some basic notions of logic programming
\cite{Apt97,Llo87} and probabilistic logic programming 
\cite{RS18}.

\subsection{Logic Programming} \label{sec:lp}

We consider a \emph{function-free}
first-order language with a fixed
vocabulary of predicate symbols, constants, and variables
denoted by $\Pi$, $\cC$ and $\cV$, respectively.
An \emph{atom} has the form $f(t_1,\ldots,t_n)$ with $f/n \in \Pi$ 
and $t_i \in (\cC\cup\cV)$ for $i = 1,\ldots,n$.  
A \emph{literal} $l$ is an atom $a$ or its negation $\neg a$.
A \emph{query} $Q$ 
is a conjunction of literals,\footnote{As is
common in logic programming, we write a query 
$l_1\wedge l_2\wedge \ldots\wedge l_n$ as
$l_1,l_2,\ldots,l_n$.
} 
where the empty query is denoted by $\Box$.
We use capital letters to denote (possibly atomic) queries.
A \emph{clause} has the form $h\leftarrow B$, 
where $h$ (the \emph{head}) is a positive literal (an atom) 
and $B$ (the \emph{body}) is a query;
when the body is empty, 
the clause is called a \emph{fact} and denoted just
by $h$; otherwise, it is called a \emph{rule}.
A (normal) logic \emph{program} $P$ is a finite set of clauses.

We let $\var(s)$ denote the set of variables in 
the syntactic object $s$,
where $s$ can be a literal, a query or a clause.  A syntactic
object $s$ is \emph{ground} if $\var(s)=\emptyset$. 
Substitutions and their operations are defined as usual,
where $\dom(\sigma) = \{x \in \cV \mid \sigma(x) \neq
x\}$ is called the \emph{domain} of a substitution $\sigma$. 
We let $\id$ denote the empty substitution.
The application of a substitution
$\theta$ to a syntactic object $s$ is usually denoted by
juxtaposition, i.e., we write $s\theta$ rather than $\theta(s)$. 
A syntactic object $s_1$ is \emph{more general} than a syntactic
object $s_2$, denoted $s_1 \sleq s_2$, if there exists a substitution
$\theta$ such that $s_1\theta = s_2$. A \emph{variable renaming} is a
substitution that is a bijection on $\cV$. 
A substitution $\theta$ is a \emph{unifier} of two syntactic objects
$s_1$ and $s_2$ iff $s_1\theta = s_2\theta$; furthermore, $\theta$ is
the \emph{most general unifier} of $s_1$ and $s_2$, denoted by
$\mgu(s_1,s_2)$ if, for every other unifier $\sigma$ of $s_1$ and
$s_2$, we have that $\theta \sleq \sigma$,\footnote{Here,
we assume that $\dom(\theta)\subseteq \var(s_1)\cup\var(s_2)$
if $\mgu(s_1,s_2)=\theta$.} i.e., there exists a substitution
$\gamma$ such that $\theta\gamma = \sigma$ when the domains are
restricted to the variables of $\var(s_1)\cup\var(s_2)$.

In this work, we consider \emph{negation as failure} \cite{Cla77} 
and SLDNF-resolution \cite{AD94}.
We say that a query $Q=l_1,\ldots,l_n$ 
\emph{resolves} to $Q'$ via $\sigma$
with respect to literal $l_i$ and clause $c$,
in symbols $Q \leadsto_\sigma Q'$, 
if either 
i) $h\leftarrow B$ is a renamed apart variant of $c$,
$\sigma = \mgu(l_i,h)$, and 
$Q' = (l_1,\ldots,l_{i-1},B,l_{i+1},\ldots,l_n)\sigma$, or
ii) $l_i$ is a negative literal, $\sigma = \id$, and 
$Q' = l_1,\ldots,l_{i-1},l_{i+1},\ldots,l_n$.
A (finite or infinite) 
sequence of resolution steps of the form
$Q_0 \leadsto_{\sigma_1} Q_1 
\leadsto_{\sigma_2} \ldots$ is called a 
\emph{pseudoderivation}. 
As we will see below, an SLDNF-derivation is a pseudoderivation
where the deletion of negative (ground) literals is justified
by a finitely failed SLDNF-tree.

An SLDNF-tree $\Gamma$ is given by a triple $(\cT,T,\subs)$, where
$\cT$ is a set of trees, $T\in\cT$ is called the main tree,
and $\subs$ is a function assigning to some nodes of trees
in $\cT$ a (subsidiary) tree from $\cT$. Intuitively speaking,
an SLDNF-tree 
is a directed graph with two types of edges, the usual ones
(associated to resolution steps) and the ones connecting a
node with the root of a subsidiary tree.
A node can be marked with \emph{failed}, \emph{sucess},
and \emph{floundered}. 
A tree is \emph{successful} if it contains at least a leaf marked
as success, and \emph{finitely failed} if it is finite and
all leaves are marked as \emph{failed}.

Given a query $Q_0$, an SLDNF-tree for $Q_0$
starts with a single node labeled with $Q_0$. 
The tree can then be
\emph{extended} by selecting a query $Q= l_1,\ldots,l_n$
which is not yet marked (as failed, success or floundered)
and a
literal $l_i$ 
and then proceeding as follows:
\begin{itemize}
\item If $l_i$ is an atom, we add a child labeled 
with $Q'$ for each resolution step 
$Q \leadsto_\sigma Q'$.
The query is marked as \emph{failed} if no such resolution
steps exist.
\item If $l_i$ is a negative literal, $\neg a$, we have the 
following possibilities:
\begin{itemize}
\item if $a$ is nonground, the query $Q$ is marked as 
\emph{floundered};
\item if $\subs(Q)$ is undefined, a new tree $T'$ with a
single node labeled with $a$ is added to $\cT$ and 
$\subs(Q)$ is set to the root of $T'$;
\item if $\subs(Q)$ is defined and the corresponding
tree is successful,
then $Q$ is marked as \emph{failed}.
\item if $\subs(Q)$ is defined and the corresponding tree is
finitely failed,
then we have $Q \leadsto_\id Q'$, 
where $Q'$ is obtained from
$Q$ by removing literal $l_i$.
\end{itemize}
\end{itemize}
Empty leaves are marked as success.
The extension of an SLDNF-tree continues until all leaves 
of the trees in $\cT$ are marked.\footnote{In \cite{AD94}
only the limit of the sequence of trees is called an 
SLDNF-tree, while the previous ones are called pre-SLDNF-trees. 
We ignore this distinction here for simplicity.}
An SLDNF-tree for a query $Q$ is an SLDNF-tree 
in which the root of the main tree is labeled with $Q$.
An SLDNF-tree is called \emph{successful} (resp.\ 
finitely failed) if the main tree is successful (resp.\
finitely failed).
An SLDNF-derivation for a query $Q$ is a branch
in the main tree of an SLDNF-tree $\Gamma$ for $Q$, together 
with the set of all trees in $\Gamma$ whose roots can
be reached from the nodes of this branch.
Given a successful SLDNF-derivation,
$Q_0 \leadsto_{\sigma_1} Q_1 \leadsto_{\sigma_2}
\ldots \leadsto_{\sigma_n} Q_n$, 
the composition $\sigma_1\sigma_2\ldots\sigma_n$ (restricted to
the variables of $Q_0$) is called a 
\emph{computed answer substitution} of $Q_0$.

A normal logic program is \emph{range-restricted} if all the variables 
occurring in the head of a clause also occur in the positive 
literals of its body. 
%
%
For range-restricted programs
every successful SLDNF-derivation 
completely grounds the initial query \cite{Mug00}. 

\subsection{Logic Programs with Annotated Disjunctions} \label{sec:plp}

We assume that $\Pi=\Pi_p\uplus\Pi_d$, 
the set of predicate symbols, is
partitioned into a set $\Pi_p$ of \emph{probabilistic predicates}
and a set $\Pi_d$ of \emph{derived predicates}, which are disjoint. 
An atom $f(t_1,\ldots,t_n)$ is called a \emph{probabilistic atom}
if $f\in\Pi_p$ and a \emph{derived atom} if $f\in\Pi_d$.
An LPAD $\cP = \cP_p\uplus\cP_d$---or just \emph{program} when
no confusion can 
arise---consists of a set of probabilistic clauses $\cP_p$
and a set of normal clauses $\cP_d$.
A \emph{probabilistic clause} has the form
$h_1\cons p_1;\ldots;h_n\cons p_n \leftarrow B$,
where $h_1,\ldots,h_n$ are atoms, $p_1,\ldots,p_n$
are real numbers in the interval $[0, 1]$ (their respective probabilities)
such that $\sum_{i=1}^n p_i \sleq 1$,
and $B$ is a query. When $\sum_{i=1}^n p_i < 1$, we 
implicitly assume that 
a special atom $\none$ is added to the head
of the clause, where $\none/0$ 
is a fresh predicate which does not occur in the
original program, with associated probability 
$1 - \sum_{i=1}^n p_i$. Thus, in the following, we
assume w.l.o.g.\ that $\sum_{i=1}^n p_i = 1$ for
all clauses.

\begin{example}  \label{ex:one}
Consider the following clause\footnote{Here and in the
remaining examples we do not show the occurrences of $\none$.}\\[1ex]
\mbox{}\hspace{4ex}$
\blue{covid(X)\cons 0.4; flu(X)\cons 0.3 \leftarrow 
  contact(X,Y), covid(Y).}
$\\[1ex]
It stats that, if $X$ is a contact of $Y$ and $Y$ has covid,
then either $X$ has covid too (probability $0.4$) or $X$
has the flu (probability 0.3). Moreover, $X$ has neither
covid nor the flu (i.e., $\none$ holds) 
with probability $0.3$ ($1-0.4-0.3$).
\end{example}
Now we consider the semantics of programs.
Given a probabilistic clause 
$c = (h_1\cons p_1;\ldots;h_n\cons p_n \leftarrow B)$, each ground
instance $c\theta$ represents a choice between $n$
(normal) clauses:
$(h_1 \leftarrow B)\theta$,
\ldots
$(h_n \leftarrow B)\theta$.
A particular choice is denoted 
by a triple $(c,\theta,i)$, $i\in\{1,\ldots,n\}$, 
which is called an \emph{atomic choice}, and has as
associated probability $\pi(c,i)$, i.e., $p_i$
in the clause above.
We say that a set $\kappa$ of atomic choices is \emph{consistent}
if it does not contain two atomic choices for the same grounding
of a probabilistic clause, i.e., it cannot contain 
$(c,\theta,i)$ and $(c,\theta,j)$ with $i\neq j$.
A set of consistent atomic choices is called a 
\emph{composite choice}. It is called a \emph{selection}
when the composite choice includes an atomic choice for
each grounding of each probabilistic clause of the program.
We let $\cS_\cP$ denote the set of all possible selections
for a given program (which is finite since we consider
function-free programs).
Each selection 
$s\in\cS_\cP$ identifies a \emph{world} $\omega_s$
which contains a
(ground) normal clause $(h_i \leftarrow B)\theta$ for each 
atomic choice $(c,\theta,i)\in s$, together with the clauses
for derived predicates:\\[1ex]
\mbox{}\hspace{4ex}$
\blue{\omega_s
= \{ (h_i \leftarrow B)\theta \mid
c=(h_1\cons p_1;\ldots;h_n\cons p_n \leftarrow B)\in\cP_p\wedge
(c,\theta,i)\in s\} \cup \cP_d}
$\\[1ex]
We assume in this work that programs are \emph{sound}, i.e., 
each world has a unique two-valued well-founded model \cite{GRS91}
which coincides with its stable model \cite{GL88},
and SLDNF-resolution is sound and complete.
We write $\omega_s\models Q$ to denote that
the (ground) query $Q$ is true in the unique model of the program.
Soundness can be ensured, e.g., 
by requiring logic programs
to be stratified \cite{Lif88}, acyclic \cite{AB91} or 
modularly acyclic \cite{Ros91}. These characterizations
can be extended to LPADs in a natural way, 
e.g., an LPAD is stratified if each
possible world is stratified.

Given a selection $s$, the probability of world $\omega_s$ is then 
defined as follows:
$
P(\omega_s) =  P(s) = \prod_{(c,\theta,i)\in s} \pi(c,i)
$.
Given a program $\cP$, we let $\cW_\cP$ denote the (finite)
set of possible worlds, 
i.e., $\cW_\cP = \{\omega_s \mid s\in\cS_\cP\}$.
Here, $P(\omega)$ defines a probability distribution over 
$\cW_\cP$.
%
By definition, the sum of the probabilities of all possible 
worlds is equal to $1$.
%
The probability of a ground query $Q$ in a program $\cP$,
called the \emph{success probability} of $Q$ in $\cP$, 
in symbols $P(Q)$, is obtained by marginalization from
the joint distribution $P(Q,\omega)$ as follows:\\[1ex]
\mbox{}\hspace{4ex}$
\blue{P(Q) = \sum_{\omega\in \cW_\cP} P(Q,\omega) = 
\sum_{\omega\in \cW_\cP} P(Q|\omega) \cdot 
P(\omega)}
$\\[1ex]
where $P(Q|\omega) = 1$ if $\omega\models Q$ 
and $P(Q|\omega) = 0$  otherwise.
Intuitively speaking, the success probability of a query is the
sum of the probabilities of all the worlds where this query
is provable (equivalently, has a successful SLDNF-derivation).

\section{Query-Driven Inference in PLP}

In this section, we present our approach to query-driven 
inference for probabilistic logic programs.
%
%
%
In order to ease the understanding, let us first 
consider the case of programs 
\emph{without negation}.
In this case, it suffices to redefine queries to also include
an associated composite choice $\kappa$. 
Intuitively speaking, $\kappa$
denotes a restriction on the worlds where the 
computation performed so far can be proved. 

An initial
query has thus the form $\tuple{Q,\emptyset}$, where $Q$ is a
standard query and $\emptyset$ is an empty composite choice. 

Resolution steps with probabilistic clauses should
update the current composite choice accordingly. 
For this operation to be well-defined,
computations must be performed w.r.t.\ the grounding $\cG(\cP)$ 
of program $\cP$ (which is finite since the considered language is 
function-free).\footnote{In practice, given a query, it suffices to compute the \emph{relevant} groundings of $\cP_p$ for 
this query (see \cite[Section 5.1]{FBRSGTJR15} for a discussion
on this topic).} 
Given a query $\tuple{Q,\kappa}$, resolution is then defined
as follows:\footnote{For simplicity, 
we use the same arrow $\leadsto$
for both standard resolution and its extended version
for probabilistic logic programs.}
\begin{itemize}
\item If the selected atom is a derived atom and
$Q\leadsto_\sigma Q'$, then $\tuple{Q,\kappa} \leadsto_{\sigma}
\tuple{Q',\kappa}$.

\item If the selected atom is a probabilistic atom, $a$,
then we have a resolution step $\tuple{Q,\kappa}
\leadsto_{\sigma} \tuple{Q',\kappa\cup\{(c,\theta,i)\}}$
for each (ground) clause $c\theta = (h_1\cons p_1;
\ldots;h_n\cons p_n\leftarrow B)\theta\in\cG(\cP)$ such that
$Q\leadsto_\sigma Q'$ is a resolution step 
with respect to atom $a$ and clause 
$h_i\theta \leftarrow B\theta$ and, moreover,
$\kappa\cup\{(c,\theta,i)\}$ is consistent.
\end{itemize}

\begin{example} \label{ex:running-pos}
  Consider the following program $\cP$:\\[1ex]
  \mbox{}\hspace{4ex}$
  \begin{array}{ll}
  \blue{(c_1)} & \blue{covid(X)\cons 0.9 \leftarrow pcr(X).}\\
  \blue{(c_2)} & \blue{covid(X)\cons 0.4; flu(X)\cons 0.3 \leftarrow 
  contact(X,Y), covid(Y).} \\ 

& \blue{pcr(p1). ~~pcr(p2). ~~contact(p1,p2). 
~~ person(p1).  ~~person(p2).  ~~person(p3).}\\

  \end{array}
  $\\[1ex]
  Here, the grounding $\cG(\cP)$ will contain an instance
  of $c_1$ and $c_2$ for each person $p1$, $p2$, and $p3$. 
  \begin{figure}[t]
  \centering
  $
  \xymatrix@C=2em@R=3em{
	& covid(p1) \ar[dl]_{\red{(c_1,\{X/p1\},1)}} \ar[dr]^{~~~\red{(c_2,\{X/p1,Y/p2\},1)}} \\
	pcr(p1) \ar[d] & & contact(p1,p2),covid(p2) \ar[d] \\
	\Box &  & covid(p2) \ar[dl]_{\red{(c_1,\{X/p2\},1)}~~} \ar[dr]^{~~~\red{(c_2,\{X/p2,Y/p3\},1)}} \\
	& pcr(p2) \ar[d] 
	& & 
    \hspace{-4ex}\txt{$\vdots$\\
    \red{[\emph{failed}]}}\\
	& \Box \\
  }
  $
  \caption{Tree for the query $covid(p1)$}
  \label{fig:ex-running-pos}
  \end{figure}
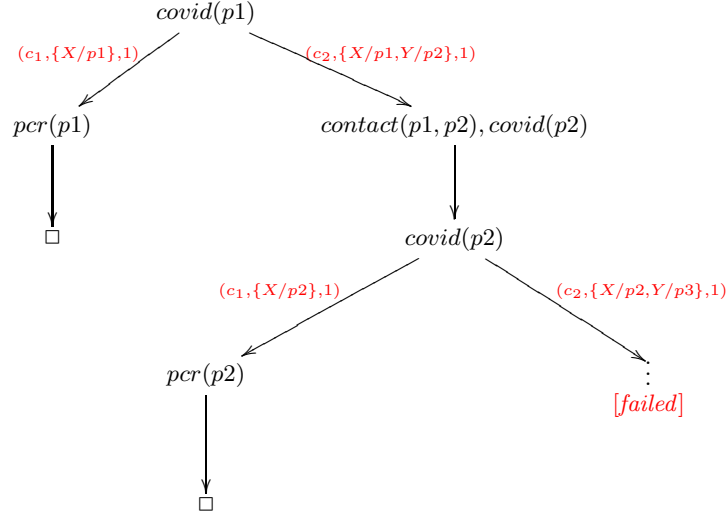
  Figure~\ref{fig:ex-running-pos} shows the resolution tree for
  the query $covid(p1)$. For simplicity, we only considered 
  two groundings for $c_2$: $\{X/p1,Y/p2\}$ and $\{X/p2,Y/p3\}$;
  moreover, we always select the leftmost literal in a query.
  In the tree, for clarity, we show ordinary queries
  as nodes and label the edges with the atomic choices computed
  in the step (if any).
  
  Here, we have two successful derivations for $covid(p1)$ that
  compute the composite choices 
  $\{(c_1,\{X/p1\},1)\}$ and $\{(c_2,\{X/p1,Y/p2\},1),
  (c_1,\{X/p2\},1)\}$, i.e., the union of the atomic choices
  labeling the steps of 
  each root-to-leaf successful derivation.
  Their probabilities are
  $\pi(c_1,1) = 0.9$ and $\pi(c_2,1)\times\pi(c_1,1) 
  = 0.4*0.9 = 0.36$, respectively. 
  The computation of the marginal probability of a query
  is not generally the sum of the probabilities of its
  proofs since the associated worlds may overlap 
  (as in this case). Computing the probability of a query
  is an orthogonal issue which is out of the 
  scope of this paper
  (in this case,
  the marginal probability is $0.936$); 
  see Section~\ref{sec:explaining} 
  for further details on this topic.
\end{example}

\subsection{Introducing Negation}

Now, we consider the general case. 
%
In the following, we say that a selection $s$ 
\emph{extends} a composite choice $\kappa$ 
if $\kappa\subseteq s$.
%
Moreover, a composite choice $\kappa$ identifies 
the set of worlds $\omega_\kappa$ that can be obtained
by extending $\kappa$ to a selection in all possible ways.
Formally, $\omega_\kappa = \{ \omega_s \mid s\in\cS_\cP\
\wedge \kappa\subseteq s \}$.
Given a set of composite choices $K$, we let
$\omega_K = \cup_{\kappa\in K}~ \omega_\kappa$.

In principle, we could adapt Riguzzi's strategy in
\cite{Rig09} for ICL (\emph{Independent Choice Logic} 
\cite{Poo97})
to the case of LPAD. Basically, a resolution step for a query 
where a negated (ground) literal $\neg a$ is selected 
could proceed as follows:
\begin{itemize}
\item First, as in SLDNF, a tree for query $a$ is built. 
Assume that the successful branches of this tree
compute the composite choices $\kappa_1,\ldots,\kappa_n$,
$n>0$.
\item Then, we know that $a$ succeeds---equivalently,
$\neg a$ fails---in all the worlds that extend 
the composite choices $\kappa_1,\ldots,\kappa_n$. 
Hence, we calculate a set of composite choices $K$
that are \emph{complementary}
to those in $\kappa_1,\ldots,\kappa_n$. 
\item If $K$ is not empty, the query where $\neg a$
was selected will have as many children as
composite choices in $K$. For each child, the negated
literal is removed from the query and the corresponding
composite choice is added to the current one (assuming it
is consistent).
\end{itemize}
In order to formalize these ideas, we first recall the notion 
of \emph{complement} \cite{Poo00}: 
If $K$ is a set of composite choices, then a \emph{complement}
of $K$ is a set $K'$ of composite choices such that
for all world $\omega\in \cW_\cP$, we have $\omega\in\omega_K$ 
iff $\omega\not\in\omega_{K'}$. 
%
%
The notion of \emph{dual} \cite{Poo00,Rig09} 
is introduced to have an 
operational definition:

\begin{definition}[dual]
  If $K$ is a set of composite choices, then composite
  choice $\kappa'$ is a \emph{dual} of $K$ if 
  for all $\kappa\in K$ there exist atomic choices
  $(c,\theta,i)\in\kappa$ and $(c,\theta,j)\in\kappa'$ 
  such that $i\neq j$. 
  A dual is \emph{minimal} if no proper subset is also a dual.
  Let $\duals(K)$ be the set of minimal duals of $K$.
\end{definition}
The set of duals is indeed a complement of a set of composite choices (cf.\ Lemma~4.8 in \cite{Poo00}).
%
%
The computation of $\duals(K)$ can be carried out 
using the notion of \emph{hitting set} \cite{Rei87}. 
Let $C$ be a collection of sets. Then, set $H$ is a 
\emph{hitting set} for $C$ if $H\subseteq \bigcup_{S\in C} S$ 
and $H\cap S \neq \emptyset$ for each $S\in C$.
In particular, we only consider hitting sets where
exactly one element of each set $S \in C$ is selected
Formally, 
\blue{$
\hits(\{S_1,\ldots,S_n\}) =
\{ \{s_1,\ldots,s_n\} \mid s_1\in S_1,\ldots,s_n\in S_n\}
$}.

In the following, given an atomic choice 
$\alpha$, we let $\ol{\alpha}$
denote the relative complement (the standard notion from
set theory) of $\{\alpha\}$ w.r.t.\
the domain of possible atomic choices for the same
(ground) clause, i.e.,\\[1ex]
\mbox{}\hspace{4ex}$
\blue{\ol{(c,\theta,i)} = \{ (c,\theta,j) \mid
c= (h_1\cons p_1;\ldots; h_n\cons p_n \leftarrow B),
~i\neq j,~j\in\{1,\ldots,n \}\}
}$\\[1ex]
We use $\alpha,\alpha',\ldots$ to denote atomic choices
and $\beta,\beta',\ldots$ for either standard atomic
choices or their complements.
Furthermore, we let 
$\ol{K} = 
\{\ol{\kappa_1},\ldots,\ol{\kappa_n}\}$
if $K = \{\kappa_1,\ldots,\kappa_n\}$,
and $\ol{\kappa} =
\ol{\alpha_1}\cup\ldots\cup\ol{\alpha_m}$
if $\kappa = \{\alpha_1,\ldots,\alpha_m\}$.
The duals of a set of composite choices $K$ can then be 
obtained from the hitting sets of $\ol{K}$:

\begin{definition}[$\duals$]
 Let $K$ be a set of composite choices. Then, 
 $
 \duals(K) =   \mins(\hits(\ol{K})) 
 $,
 where function $\mins$ removes inconsistent and 
 redundant composite choices, i.e., 
$\mins(K) = \{\kappa\in K\mid consistent(\kappa) ~\mbox{and}~
\kappa' \not\subset \kappa~\mbox{for all}~\kappa'\in K\}$.
\end{definition}
It is easy to see that the above definition is more declarative
but equivalent to similar functions in \cite{Poo00,Rig09}.
%

\begin{example} \label{ex:running-neg}
  Consider the following LPAD program $\cP$:\\[1ex]
  \mbox{}\hspace{4ex}\blue{$
  \begin{array}{ll}
  (c_1) & covid(X)\cons 0.9 \leftarrow pcr(X).\\
  (c_2) & covid(X)\cons 0.4; flu(X)\cons 0.3 \leftarrow 
  contact(X,Y), covid(Y), \neg protected(X).\\
  (c_3) & \ffp(X)\cons 0.3; surgical\cons 0.4; cloth\cons 0.1 \leftarrow person(X).\\
  (c_4) & vaccinated(X)\cons 0.8 \leftarrow person(X).\\
  (c_5) & vulnerable(X)\cons 0.6 \leftarrow \neg young(X).\\
  (c_6) & young(X)\cons 0.2; adult(X)\cons 0.5 \leftarrow person(X).\\[1ex]

& protected(X) \leftarrow \ffp(X).\\
& protected(X) \leftarrow vaccinated(X), \neg vulnerable(X).\\
& pcr(p1). ~~pcr(p2). ~~contact(p1,p2). 
~~ person(p1).  ~~person(p2).  ~~person(p3).\\

  \end{array}
  $}\\[1ex]
  The grounding $\cG(\cP)$ will contain an instance
  of clauses ${c_1},\ldots,{c_6}$ for each person 
  ${p1}$, ${p2}$, 
  and ${p3}$. 
  \begin{figure}[t]
  \centering
  $
  \xymatrix@C=0ex@R=2em{
  & {pro(p1)} \ar[dl] \ar[dr] &&&& 
  {vuln(p1)} \ar[d]^{\red{(c_5,\sigma,1)}} 
  &&&& {young(p1)} \ar[d]_{\red{(c_6,\sigma,1)}}\\  
  {\ffp(p1)} \ar[d]_{\red{(c_3,\sigma,1)}} && {vacc(p1),\neg vuln(p1)} 
  \ar[d]^{\red{(c_4,\sigma,1)}} &&& {\neg young(p1)} \ar[dl]_{\red{(c_6,\sigma,2)}} 
  \ar[dr]^{\red{(c_6,\sigma,3)}} \ar@{.>}@/^/[rrrru]
  &&&& {person(p1)} \ar[d] 
  \\  
  {person(p1)} && {person(p1),\neg vuln(p1)} 
  \ar[d] &&\Box && \Box &&& \Box\\
  && {\neg vuln(p1)} \ar[dl]_{\red{(c_5,\sigma,2)}} \ar[dr]^{\red{(c_6,\sigma,1)}} \ar@{.>}@/^/[rrruuu] \\
  & \Box && \Box 
  }
  $ 
  \caption{Trees for the query {$protected(p1)$}, where 
  predicates 
  {$protected$}, {$vaccinated$}, and 
  {$vulnerable$} are abbreviated as
  {$pro$}, {$vacc$}, and {$vuln$}, respectively. }
  \label{fig:ex-running-pro}
  \end{figure}
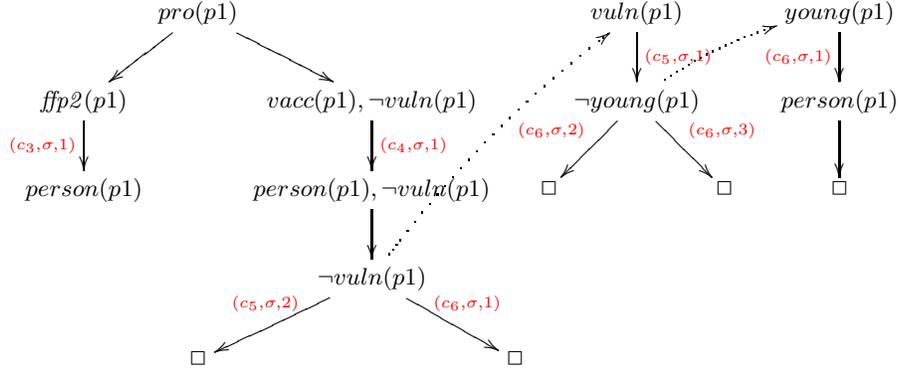
  As in the previous example,
  we show ordinary queries
  as nodes and label the edges with the new
  atomic choices of the step (if any).
  %
  Figure~\ref{fig:ex-running-pro} shows the trees for
  the query {$protected(p1)$}.
  Here, {$young(p1)$} has only one successful
  derivation with composite choice 
  {$\{(c_6,\sigma,1)\}$}, where {$\sigma = \{X/p1\}$}.
  Let {$K_1 =\{\{(c_6,\sigma,1)\}\}$}.
  In order to resolve {$\neg young(p1)$}, we compute
  the duals of {$K_1$}:\\[1ex]
  \mbox{}\hspace{4ex}\blue{$
  \begin{array}{lll}
  \duals(K_1)
  & = & \mins(\hits(\ol{K_1})) 
  = \mins(\hits(\{\ol{\{(c_6,\sigma,1)\}}\})) \\
  & = & \mins(\hits(\{\{(c_6,\sigma,2),(c_6,\sigma,3)\}\}))
   =  \{\{ (c_6,\sigma,2)\},\{(c_6,\sigma,3)\}\}\\
  \end{array}
  $}\\[1ex]
  Hence, {$\neg young(p1)$} has two children with
  atomic choices {$(c_6,\sigma,2)$} and
  {$(c_6,\sigma,3)$}.
  Consider now {$vulnerable(p1)$}. The
  tree includes two successful branches that compute
  composite choices {$\{(c_5,\sigma,1),(c_6,\sigma,2)\}$} and
  {$\{(c_5,\sigma,1),(c_6,\sigma,3)\}$}. Let {$K_2$} 
  be a set
  with these two composite choices.
  In order to resolve {$\neg vulnerable(p1)$} 
  in the tree for 
  {$protected(p1)$}, we first need the duals of 
  {$K_2$}:\\[1ex]
  \mbox{}\hspace{4ex}\blue{$
  \begin{array}{lll}
    \duals(K_2) =  \mins(\hits(\ol{K_2})) \\
   =  \mins(\hits(\{\ol{\{(c_5,\sigma,1),(c_6,\sigma,2)\}},
  \ol{\{(c_5,\sigma,1),(c_6,\sigma,3)\}}\})) \\
  = \mins(\hits(\{\{(c_5,\sigma,2),(c_6,\sigma,1),(c_6,\sigma,3)\},
  \{(c_5,\sigma,2),(c_6,\sigma,1),(c_6,\sigma,2)\}\})) \\
  = \{\{(c_5,\sigma,2)\},
  \{(c_6,\sigma,1)\}\}
  \end{array}
  $}\\[1ex]
  Note that function {$\hits$} compute several other sets 
  of atomic choices but they
  are discarded by function {$\mins$} either because they are
  inconsistent (the case of 
  {$\{(c_6,\sigma,2),(c_6,\sigma,3)\}$})
  or because they are redundant since they are supersets 
  of {$\{(c_5,\sigma,2)\}$} or {$\{(c_6,\sigma,1)\}$}. 
  Hence, {$\neg vulnerable(p1)$} has two children with
  associated atomic choices {$(c_5,\sigma,2)$} and
  {$(c_6,\sigma,1)$}.
    
  We do not show the details here but, given
  the computed composite choices
  {$\{(c_3,\sigma,1)\}$}, 
  {$\{(c_4,\sigma,1),(c_5,\sigma,2)\}$},
  and {$\{(c_4,\sigma,1),(c_6,\sigma,1)\}$} for
  {$protected(p1)$}, a call of the
  form {$\neg protected(p1)$} would
  have twelve children.
\end{example}

\subsection{An Algebra of Choice Expressions}

The main drawback of the previous approach is that it
usually produces a large number of proofs (i.e., 
successful branches), most of them identical except for
the computed composite choice. 
For instance, as mentioned in Example~\ref{ex:running-neg},
a call to $\neg protected(p1)$ will have twelve children.
Likewise, a query of the form 
$\neg protected(p1),\neg protected(p2)$ will end up with
a total of 144 children, all of them with a copy of
the same query.

Our focus in this work is explainability. Hence, we aim 
at producing proofs that are as simple and easy to understand
and manipulate as possible. 
An obvious first step into this direction could consist in
associating \emph{a set of composite choices} to each query rather
than a single composite choice. 
In this way, 
the resolution of a query with a negative literal would 
produce a single child with the
composition of the current set of composite choices and
those returned by function $\duals$.

E.g., the query $\neg vulnerable(p1)$ in the first
tree of Figure~\ref{fig:ex-running-neg} would have the
associated set $K = \{\{(c_4,\sigma,1)\}\}$. 
Then, given the output of $\duals(K_2)$ shown above, 
the child of $\neg vulnerable(p1)$ would have the following
associated set of composite choices:
$
K\otimes \duals(K_2) = 
\{\{(c_4,\sigma,1),(c_5,\sigma,2)\},
  \{(c_4,\sigma,1),(c_6,\sigma,1)\}\}
$,
where the operation ``$\otimes$'' is defined as follows:
\begin{equation} \label{eqn:otimes}
\blue{K_1 \otimes K_2 = \mins(\{
\kappa_1 \cup \kappa_2 \mid \kappa_1\in K_1, \kappa_2\in K_2\})}
\end{equation}
However, this approach would only reduce the number of 
identical nodes
in the tree, but the computed composite choices
would be the same as before.

As an alternative, we introduce an algebra of
\emph{choice expressions},
a representation for sets of composite choices 
which enjoy several good properties:
they are more compact and can be easily manipulated 
using well-known logical rules (distributive laws, 
double negation elimination, De Morgan's laws, etc). 

\begin{definition}[choice expression]
  A {choice expression} is defined inductively as follows:
  \begin{itemize}
  \item $\bot$ and $\top$ are choice expressions;
  \item an atomic choice $\alpha$ is a choice expression;
  \item if $\ch,\ch'$ are choice expressions then 
  $\neg \ch$, $\ch\wedge\ch'$, and $\ch\vee\ch'$ are 
  choice expressions too, where 
  $\neg$ has higher precedence than $\wedge$, and 
  $\wedge$ higher than $\vee$.
  \end{itemize}
  Given a program $\cP$, we let $\bigch_\cP$ denote the
  associated domain of choice expressions that can be built
  using the atomic choices of $\cP$.
  We will omit the
  subscript $\cP$ in $\bigch_\cP$
  when the program is clear from the 
  context or irrelevant.
\end{definition}
A choice expression essentially represents a set
of composite choices. 
For example, the expression
$(\alpha_1\wedge\alpha_2)\vee 
\alpha_3\vee(\alpha_4\wedge\alpha_5)$ represents
the set $\{\{\alpha_1,\alpha_2\},\{
\alpha_3\},\{\alpha_4,\alpha_5\}\}$.
In particular, negation allows us to represent sets of 
composite choices in a much more compact way. 
E.g., $\neg(c_3,\sigma,1)\wedge\neg(c_6,\sigma,1)$ represents 
a set with 6 composite choices with all combinations 
of pairs from $\{(c_3,\sigma, 2),(c_3,\sigma,3),(c_3,\sigma,4)\}$
and $\{(c_6,\sigma,2),(c_6,\sigma,3)\}$.
The following function $\gamma$ formalizes this equivalence:

\begin{definition}
  Given a choice expression $\ch\in\bigch$, we let $\gamma(\ch)$
  denote the associated set of composite choices,
  where function $\gamma$ is defined inductively as follows:
  \begin{itemize}
  \item $\gamma(\bot) = \{\}$, i.e., $\bot$ denotes an
  inconsistent set of atomic choices.
  \item $\gamma(\top) = \{\{\}\}$, i.e., $\top$ represents
  a composite choice, $\{\}$, that can be extended in order
  to produce all possible selections. 
  \item $\gamma(\alpha) = \{\{\alpha\}\}$.
  \item $\gamma(\neg \ch) = \duals(\gamma(\ch))$, i.e., $\neg \ch$
  represents a complement of $\ch$.
  \item $\gamma(\ch_1\wedge\ch_2) = \mins(\gamma(\ch_1) \otimes
  \gamma(\ch_2))$, where ``\:$\otimes$\!'' is defined 
  in (\ref{eqn:otimes}) above.
  \item $\gamma(\ch_1\vee\ch_2) =
  \mins(\gamma(\ch_1)\cup\gamma(\ch_2))$.
  \end{itemize}
\end{definition}
In practice, a set of composite choices $K$ is just a 
device to represent a set of worlds $\omega_K$. Therefore,
we will not distinguish two choice expressions, $\ch_1$ and
$\ch_2$, as long as the worlds identified by $\gamma(\ch_1)$
and $\gamma(\ch_2)$ are the same. 
For example, the choice expressions $\alpha_1$
and $(\alpha_1\wedge\alpha_2)\vee(\alpha_1\wedge\neg \alpha_2)$
are equivalent, since both represent the same set of 
selections (those including atomic choice $\alpha_1$).

Formally, we introduce the following equivalence relation on 
choice expressions: $\ch_1\sim\ch_2$ if 
$\omega_{\gamma(\ch_1)} = \omega_{\gamma(\ch_2)}$.
Roughly speaking, $\ch_1$ and $\ch_2$ are equivalent
if the sets of composite choices in $\gamma(\ch_1)$ and 
$\gamma(\ch_2)$ can be extended to produce the same set of 
selections. 

Let $\widetilde{\bigch}$ denote the quotient set of
$\bigch$ by ``$\sim$''. 
Moreover, we let $\ch\in \widetilde{\bigch}$
denote the equivalence class $[\ch]$ 
when no confusion can arise.
Then, $\tuple{\widetilde{\bigch},\wedge,\vee,\neg,\top,\bot}$ 
is a Boolean algebra and the following axioms hold (see
proofs in the Appendix):
\begin{description}
\item[Associativity] 
\blue{$\ch_1\vee (\ch_2\vee \ch_3) = (\ch_1\vee \ch_2) \vee \ch_3$ }
and
\blue{$\ch_1\wedge(\ch_2\wedge \ch_3) = (\ch_1\wedge \ch_2) \wedge \ch_3$}.

\item[Commutativity] \blue{$\ch_1\vee \ch_2= \ch_2\vee \ch_1$} and 
\blue{$\ch_1\wedge \ch_2 = \ch_2 \wedge \ch_1$}.

\item[Absorption] \blue{$\ch_1 \vee (\ch_1\wedge \ch_2) = \ch_1$} 
and \blue{$\ch_1 \wedge (\ch_1 \vee \ch_2) = \ch_1$}.

\item[Identity] \blue{$\ch\vee \bot = \ch$} and 
\blue{$\ch\wedge \top = \ch$}.

\item[Distributivity] \blue{$\ch_1\vee (\ch_2 \wedge \ch_3) =
(\ch_1\vee \ch_2) \wedge (\ch_1\vee \ch_3)$} and 
\blue{$\ch_1 \wedge (\ch_2\vee \ch_3) = 
(\ch_1\wedge \ch_2)\vee (\ch_1\wedge \ch_3)$}.

\item[Complements] \blue{$\ch\vee \neg \ch = \top$} and 
\blue{$\ch\wedge \neg \ch = \bot$}.
\end{description}
Furthermore, double negation elimination and De Morgan's laws
also hold:
\begin{description}
\item[Double negation elimination] \blue{$\neg\neg\ch = \ch$}.
\item[De Morgan] \blue{$\neg(\ch_1\vee\ch_2) = \neg\ch_1\wedge\neg\ch_2$}
and \blue{$\neg(\ch_1\wedge\ch_2) = \neg\ch_1\vee\neg\ch_2$}.
\end{description}
In the following, function ``$\mins$'' is 
redefined in terms of the 
following rewrite rules which are applied modulo
associativity and commutativity of $\wedge$ and $\vee$:\\[1ex]
\mbox{}\hspace{4ex}$
\blue{\begin{array}{rcl@{~~}l@{~~~~~~}rcl@{~~~~~~}rcl}
  \alpha_1\wedge\alpha_2 & \to & \bot 
  & \mbox{if}~\alpha_1\in\ol{\alpha_2}
  &   \ch\wedge\neg\ch & \to & \bot
  & \ch\wedge \top & \to & \ch \\
  
  \alpha_1\wedge\neg\alpha_2 & \to & \alpha_1
  & \mbox{if}~\alpha_1\in\ol{\alpha_2}
  & \ch \wedge \bot & \to & \bot
  & \ch\vee\top & \to & \top\\
  
  &&&&\ch_1\vee(\ch_1\wedge\ch_2) & \to & \ch_1
  & \ch\wedge\ch & \to & \ch
\end{array}}
$\\[1ex]
The first rule introduces an inconsistency when a conjunction
includes two different atomic choices for the same clause 
$c\theta$. 
The second rule simplifies a conjunction since
$\neg \alpha_2$ denotes any atomic choice in $\ol{\alpha_2}$
but $\alpha_1\in\ol{\alpha_2}$ is more specific.
The remaining rules are just oriented axioms or an
obvious simplification.

It is often useful to compute the DNF (Disjunctive Normal Form) 
of a choice expression in order to have a canonical representation:

\begin{definition}[DNF]
  Let $\ch$ be a choice expression. Then, $\dnf(\ch)$ is
  defined as follows:
  $
  \dnf(\ch) = \mins(\ch')
  $,
  where $\ch \to^\ast \ch' \not\to$
  and the relation $\to$ is defined by the following canonical
  term rewrite system:\\[1ex]
  $
  \mbox{}\hspace{4ex}\blue{\begin{array}{rcl@{~~~~~~}rcl@{~~~~~~}rcl}
  \neg \neg \ch & \to & \ch &

  \neg (\ch_1\vee\ch_2) & \to & \neg \ch_1 \wedge \neg \ch_2
  &
    \ch_1\wedge(\ch_2\vee\ch_3) & \to & 
     (\ch_1\wedge\ch_2)\vee(\ch_1\wedge\ch_3)\\

  &&& \neg (\ch_1\wedge\ch_2) & \to & \neg \ch_1 \vee \neg \ch_2
  & (\ch_1\vee\ch_2)\wedge\ch_3 & \to & 
     (\ch_1\wedge\ch_3)\vee(\ch_2\wedge\ch_3)\\
  \end{array}}
  $
\end{definition}
In the following, if a tree for atom $a$ has $n$ successful
derivations computing choice expressions $\ch_1,\ldots,\ch_n$,
we let $\dnf(\neg(\ch_1\vee\ldots\vee\ch_n))$ denote its
duals.

\subsection{\sldp-Resolution}

Finally, we can formalize the construction of 
\sldp-trees.\!\footnote{\sldp\ stands for Selection 
rule driven Linear resolution for Probabilistic
Definite clauses augmented by the Negation as Failure rule.}  
In principle,
they have the same structure of SLDNF-trees. The main
difference is that nodes are now labeled with pairs
$\tuple{Q,\ch}$ and that the edges are labeled with
both an mgu (as before) and a choice expression (when a 
probabilistic atom is selected). 
Given a query $Q_0$, the construction of an 
\sldp-tree for $Q_0$
starts with a single node labeled with $\tuple{Q_0,\top}$. 
An \sldp-tree can then be
\emph{extended} by selecting a query $\tuple{Q,\ch}$
with $Q= l_1,\ldots,l_n$
which is not yet marked (as failed, success or floundered)
and a
literal $l_i$ of $Q$ 
and then proceeding as follows:
\begin{itemize}
\item If $l_i$ is a derived atom and
$Q\leadsto_\sigma Q'$, then $\tuple{Q,\ch} \leadsto_{\sigma}
\tuple{Q',\ch}$.

\item If $l_i$ is a probabilistic atom,
then we have a resolution step $\tuple{Q,\ch}
\leadsto_{\sigma,(c,\theta,i)} \tuple{Q',\ch'}$
for each clause $c\theta = (h_1\cons p_1;
\ldots;h_n\cons p_n\leftarrow B)\theta\in\cG(\cP)$ such that
$Q\leadsto_\sigma Q'$ is a resolution step 
with respect to atom $l_i$ and clause 
$h_i\theta \leftarrow B\theta$ and, moreover,
$\ch' = \dnf(\ch\wedge (c,\theta,i)) \neq \bot$. 

\item If $l_i$ is a negative literal $\neg a$ we have the 
following possibilities:
\begin{itemize}
\item if $a$ is nonground, the query $Q$ is marked as 
\emph{floundered};
\item if $\subs(Q)$ is undefined, a new tree $T'$ with a
single node labeled with $\tuple{a,\top}$ is added to $\cT$ and 
$\subs(Q)$ is set to the root of $T'$;
\item if $\subs(Q)$ is defined, the corresponding
tree cannot be further extended, and 
it has $n$ leaves marked as success with associated choice
expressions $\ch_1,\ldots,\ch_n$, $n\geq 0$, 
then we have $\tuple{Q,\ch} 
\leadsto_{\id,\ch_a} 
\tuple{Q',\ch'}$, where $Q'$ is obtained from 
$Q$ by removing literal $l_i$, $\ch_a = \dnf(\neg(\ch_1\vee\ldots\vee\ch_n))$, 
and $\ch' = \dnf(\ch\wedge \ch_a)\neq\bot$.
If $\ch'=\bot$ or $n=0$, the node is marked as failed.
\end{itemize}
\end{itemize}
Leaves with empty queries are marked as success.
An \sldp-tree for a query $Q$ is an \sldp-tree 
in which the root of the main tree is labeled 
with $\tuple{Q,\top}$.
An \sldp-tree is called \emph{successful} (resp.\ 
finitely failed) if the main tree is successful (resp.\
finitely failed).
An \sldp-derivation for a query $Q$ is a branch
in the main tree of an \sldp-tree $\Gamma$ for $Q$, together 
with the set of all trees in $\Gamma$ whose roots can
be reached from the nodes of this branch.

Given a successful \sldp-derivation for $Q$---also called 
a \emph{proof}---of the form
$\tuple{Q,\top} = \tuple{Q_0,\ch_0} \leadsto_{\sigma_1} \tuple{Q_1,\ch_1} 
\leadsto_{\sigma_2} \ldots \leadsto_{\sigma_n} \tuple{Q_n,\ch_n}
= \tuple{\Box,\ch}$, 
the composition $\sigma_1\sigma_2\ldots\sigma_n$ (restricted to
$\var(Q)$) is called a 
\emph{computed answer substitution} of $Q$
and $\ch$ represents the 
worlds where this derivation can be proved.

\begin{example} \label{ex:running-neg-2}
  Consider again the LPAD from Example~\ref{ex:running-neg}
  and the same grounding for $p1$, $p2$, and $p3$.
  As in Example~\ref{ex:running-pos}, we only consider 
  two groundings for $c_2$ for simplicity: 
  $\{X/p1,Y/p2\}$ and $\{X/p2,Y/p3\}$.
  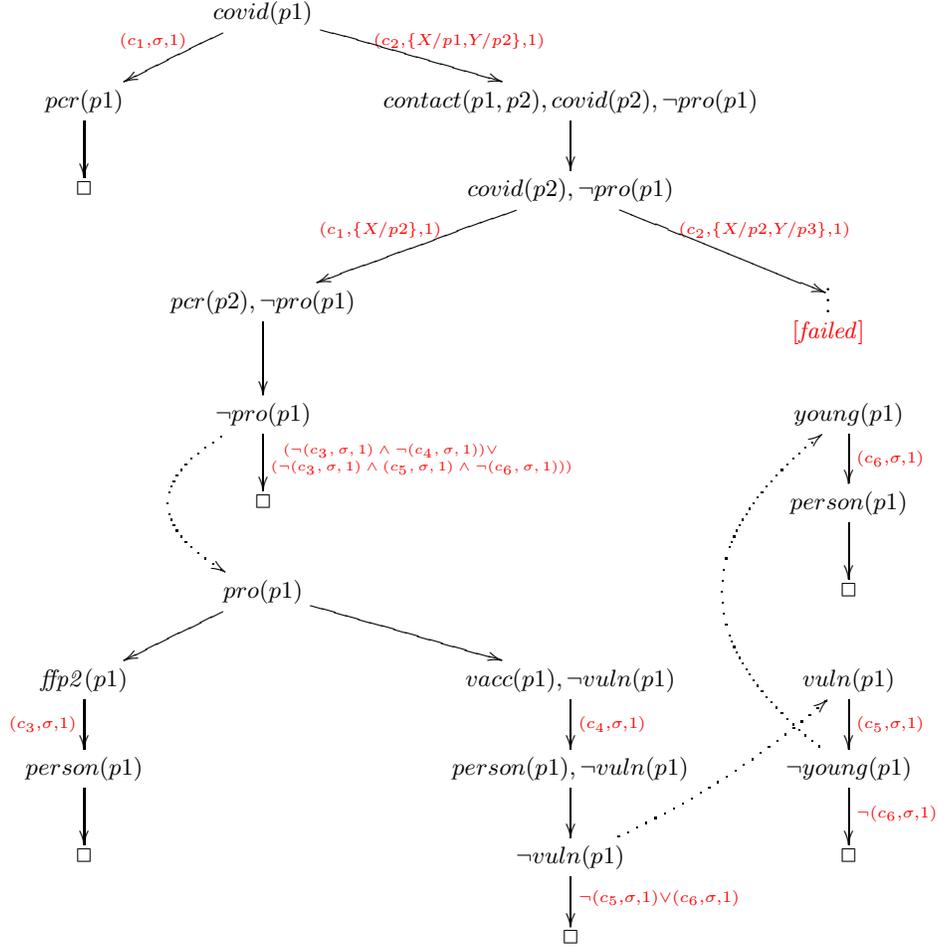
\begin{figure}[t]
  \centering
  $
  \xymatrix@C=0.5em@R=2em{
	& covid(p1) \ar[dl]_{\red{(c_1,\sigma,1)}} 
	\ar[dr]^{~~~\red{(c_2,\{X/p1,Y/p2\},1)}} \\
	pcr(p1) \ar[d] & & contact(p1,p2),covid(p2),\neg pro(p1) \ar[d] \\
	\Box &  & covid(p2),\neg pro(p1) \ar[dl]_{\red{(c_1,\{X/p2\},1)}~~} \ar[dr]^{~~~\red{(c_2,\{X/p2,Y/p3\},1)}} \\ 
	& pcr(p2),\neg pro(p1) \ar[d] & & \hspace{-4ex}\txt{$\vdots$\\
    \red{[\emph{failed}]}}\\ 
	& \neg pro(p1) \ar@{.>}@/_3pc/[dd] 
	\ar[d]^{\tiny \txt{\red{$(\neg(c_3,\sigma,1)\wedge \neg (c_4,\sigma,1)) \vee$}~~~~~~~~~~ \\ \red{$(\neg(c_3,\sigma,1)\wedge(c_5,\sigma,1)\wedge\neg (c_6,\sigma,1)))$}}} 
	& & young(p1) \ar[d]^{\red{(c_6,\sigma,1)}} \\
	& \Box  & & person(p1) \ar[d]\\
  & pro(p1) \ar[dl] \ar[dr]  & & \Box & \\
  \ffp(p1) \ar[d]_{\red{(c_3,\sigma,1)}} & 
  & vacc(p1),\neg vuln(p1)\ar[d]^{\red{(c_4,\sigma,1)}} 
  & vuln(p1) \ar[d]^{\red{(c_5,\sigma,1)}} \\
  person(p1) \ar[d] && person(p1),\neg vuln(p1)\ar[d] & \neg young(p1) \ar[d]^{\red{\neg (c_6,\sigma,1)}} \ar@{.>}@/^4pc/[uuuu] \\
  \Box && \neg vuln(p1) \ar@{.>}@/_/[ruu]
  \ar[d]^{\red{\neg (c_5,\sigma,1)\vee(c_6,\sigma,1)}} & \Box  \\
  && \Box && \\
  }
  $
  \caption{\sldp-tree for query $covid(p1)$, 
  where $\sigma = \{X/p1\}$
  and predicates 
  {$protected$}, {$vaccinated$}, and 
  {$vulnerable$} are abbreviated as
  {$pro$}, {$vacc$}, and {$vuln$}, respectively.}
  \label{fig:ex-running-neg}
  \end{figure}
  Figure~\ref{fig:ex-running-neg} shows the \sldp-tree for
  the query $covid(p1)$. 
  %
  %
  Here, we have two proofs for $covid(p1)$.
  The choice expression $\ch_1$ of the first proof is 
  just \blue{$(c_1,\sigma,1)$}, where $\sigma = \{X/p1\}$. 
  The choice expression $\ch_2$ of the second proof is given
  by\\[1ex]
  \blue{$
  \mbox{}\hspace{2ex}\begin{array}{ll}
  & (c_2,\{X/p1,Y/p2\},1)\wedge (c_1,\{X/p2\},1)\wedge
  \neg(c_3,\sigma,1)\wedge \neg (c_4,\sigma,1)\\
  \vee &
  (c_2,\{X/p1,Y/p2\},1)\wedge (c_1,\{X/p2\},1)\wedge
  \neg(c_3,\sigma,1)\wedge(c_5,\sigma,1)\wedge\neg (c_6,\sigma,1)
  \end{array}
  $}\\[1ex]
  Here, $\ch_2$ represents 
  a total of 9 composite choices which are obtained 
  by replacing negated 
  atomic choices by their corresponding atomic choices
  (i.e., $\gamma(\ch_2)$).
\end{example}
Traditionally, an explanation 
for a ground query $Q$ is defined as a selection $s$
such that $Q$ is true in the world $\omega_s$
associated to this selection.
As in \cite{Poo00}, we consider in this work a more relaxed 
notion and say that a composite choice $\kappa$ is an
explanation for ground query $Q$ if $Q$ is true in \emph{all}
worlds associated to every selection that extends $\kappa$.
Formally, $\kappa$ is an \emph{explanation} for $Q$ if
$\omega_s\models Q$ for all selection $s\supseteq \kappa$.
We also say that
a set of composite choices $K$ is \emph{covering} w.r.t.\
(ground) query $Q$ if for all $\omega\in \cW_\cP$ such that
$\omega\models Q$ we have $\omega\in\omega_K$ \cite{Poo00}.

Finding the most likely explanation of a query attracted
considerable interest in the probabilistic logic programming
field (where it is also called \emph{Viterbi proof} \cite{Poo93b}).
Note that, although it may seem counterintuitive, the 
selection with the highest probability cannot always
be obtained by extending the most likely 
proof of a query (see \cite[Example 6]{SRVKMJ14}).
%
%
Let $\expl_\cP(Q)$ denote the set of composite choices 
represented by the choice expressions
in the successful leaves of the \sldp-tree for $Q$ w.r.t.\
$\cG(\cP)$,
i.e., $\expl_\cP(Q) = \bigcup_{\ch \in L} \gamma(\ch)$,
where $L$ is the set of leaves marked
as success in the main \sldp-tree. 
The following result assumes that the query
$Q$ is ground, but could be extended to non-ground queries
by considering each proof separately and backpropagating
the computed answer substitution (which grounds the query
since the program is range-restricted \cite{Mug00}).
%

\begin{theorem} \label{th:sound}
  Let $\cP$ be a sound program and $Q$ a ground query.  
  Then, $\omega_s \models Q$ iff there exists a composite
  choice $\kappa\in\expl_\cP(Q)$ such that $\kappa\subseteq s$.
\end{theorem}
Therefore, $\expl(Q)$ is indeed a finite set of explanations
which is covering for $Q$.

\section{Proofs as Explanations} \label{sec:explaining}

In this section, we focus on the representation of the
explanations of a query.
In principle, we propose to show the proofs of a query 
(its successful \sldp-derivations)
as its explanations, from highest to lowest probability.
Previous approaches only considered the 
probability of a standard derivation $D$ computing a composite
choice $\kappa$ so that 
$P(D) = \prod_{(c,\theta,i)\in\kappa} \pi(c,i)$.
Unfortunately, this is not applicable to 
\sldp-derivations computing a choice expression
(equivalently, computing a set of composite choices).
Therefore, 
we define the probability of an \sldp-derivation as follows:

\begin{definition}
  Let $\cP$ be a program. Given a successful \sldp-derivation
  $D$ for a query $Q$ computing the choice expression $\ch$, its
  associated probability is 
  $P(D) = \sum_{\omega\in \omega_{\gamma(\ch)}}P(\omega)
  $, i.e., the sum of the probabilities of all the worlds
  where the successful derivation can be proved.
\end{definition}
Computing $P(D)$ resembles
the problem of computing the probability of a query:
summing up the probabilities of the
composite choices in $\gamma(\ch)$ for a computed 
choice expression $\ch$ would not be correct
since their associated worlds may overlap
(i.e., there might be $\kappa,\kappa'\in\gamma(\ch)$
with $\kappa\neq\kappa'$ such that 
$\omega_\kappa\cap\omega_{\kappa'}\neq\emptyset$).

There is ample literature on computing the probability of
a query, e.g., by combining inference and a conversion 
to some kind of Boolean formula \cite{FBRSGTJR15}. We consider this
problem 
an orthogonal topic which is outside of the scope of this paper.
Nevertheless, we present a transformational approach that
converts the problem of computing the probability of an 
\sldp-derivation into the problem of computing the probability
of a query in an LPAD program.

\begin{definition}
  The first transformation takes an LPAD and returns 
  a new LPAD:\\[1ex]
  \mbox{}\hspace{4ex}
  $  \blue{\begin{array}{l@{~}l}
  \trp(\cP)  = \{ ch_1\cons p_1;\ldots; ch_n\cons p_n
  & \mid c\theta = (h_1\cons p_1;\ldots;h_n\cons p_n \leftarrow B)\theta\in \cG(\cP)\\
  & ~~\mbox{and}~ch_i=ch(c,\ol{\var}(c)\theta,i),~i\in\{1,\ldots,n\}\}
  \end{array}}
  $\\[1ex]
  where $\ol{\var}(c)$ returns a \emph{list} 
  with the clause variables.
  Our second transformation takes a choice expression and
  returns a (ground) query as follows:\\[1ex]
  \mbox{}\hspace{4ex}$
  \blue{\begin{array}{l@{~~~~~}l}
    \trc(\top) = \mathit{true} &
    \trc(\bot) = \mathit{false}\\
    \trc((c,\theta,i)) = ch(c,\ol{\var}(c)\theta,i) &
    \trc(\neg \ch) = \neg \trc(\ch)\\
    \trc(\ch_1\wedge\ch_2) = \trc(\ch_1),\trc(\ch_2) &
    \trc(\ch_1\vee\ch_2) = \trc(\ch_1);\trc(\ch_2)
  \end{array}}  
  $
\end{definition}
Now, given an LPAD $\cP$, 
the probability of an \sldp-derivation $D$ computing
choice expression $\ch$ can be obtained from the
probability of query $\trc(\ch)$ in LPAD $\trp(\cP)$.
The correctness of the transformation is an easy
consequence of the fact that
the probability distribution of $\cP$ and $\trp(\cP)$
is the same and that the query $\trc(\ch)$ 
computes an equivalent choice expression in $\trp(\cP)$;
namely, an atomic choice $(c,\theta,i)$ has now
the form $(c\theta,\{\},i)$ but the structure of the 
computed choice expression is the same.

\begin{example}
  Consider again LPAD $\cP$ from Example~\ref{ex:running-neg}
  and its grounding for $p1$, $p2$, and $p3$. $\trp(\cP)$
  is as follows (for clarity, only the clauses of interest are shown):\\[1ex]
  \mbox{}\hspace{4ex}$
  \blue{\begin{array}{l}
  ch(c_1,[p1],1)\cons 0.9.~~~~~
  ch(c_1,[p2],1)\cons 0.9.\\
  ch(c_2,[p1,p2],1)\cons 0.4; ch(c_2,[p1,p2],2)\cons 0.3.\\
  ch(c_3,[p1],1)\cons 0.3; ch(c_3,[p1],2)\cons 0.4; ch(c_3,[p1],3)\cons 0.1.\\
  ch(c_4,[p1],1)\cons 0.8.~~~~~
  ch(c_5,[p1],1)\cons 0.6.~~~~~
  ch(c_6,[p1],1)\cons 0.2; ch(c_6,[p1],2)\cons 0.5.\\
  \end{array}}
  $\\[1ex]
  The query $covid(p1)$ computes two choice expressions:
  $\ch_1 = (c_1,\sigma,1)$, where $\sigma = \{X/p1\}$, and
  $\ch_2 =$\\[-3ex]
  \begin{equation} 
  \label{eqn:c2}
  \mbox{}\hspace{-3ex}\begin{array}{ll}
  & (c_2,\{X/p1,Y/p2\},1)\wedge (c_1,\{X/p2\},1)\wedge
  \neg(c_3,\sigma,1)\wedge \neg (c_4,\sigma,1)\\
  \vee &
  (c_2,\{X/p1,Y/p2\},1)\wedge (c_1,\{X/p2\},1)\wedge
  \neg(c_3,\sigma,1)\wedge(c_5,\sigma,1)\wedge\neg (c_6,\sigma,1)
  \end{array}
  \\[1ex]
  \end{equation}
  Here, we have 
  $\trc(\ch_1) = ch(c_1,[p1],1)$ and
  $\trc(\ch_2) =$\\[1ex]
  \mbox{}\hspace{4ex}$
  \blue{\begin{array}{l}
  ch(c_2,[p1,p2],1),ch(c_1,[p2],1),
  \neg ch(c_3,[p1],1),\neg ch(c_4,[p1],1)\\
   ;
  ch(c_2,[p1,p2],1), ch(c_1,[p2],1),
  \neg ch(c_3,[p1],1),ch(c_5,[p1],1),\neg ch(c_6,[p1],1)
  \end{array}}
  $\\[1ex]
  The probability of $\trc(\ch_1)$ in $\trp(\cP)$ using
  a system like PITA \cite{RS13} or ProbLog \cite{FBRSGTJR15}
  is $0.9$, while that of $\trc(\ch_2)$ is 
  $0.147168$. The probability of the query $covid(p1)$ can  
  be obtained from the disjunction $\trc(\ch_1);\trc(\ch_2)$,
  which returns $0.9147168$.
\end{example}
As mentioned before, we propose to show the proofs of a query
as explanations, each one with its associated probability.
However, instead of using \sldp-derivations, each proof will 
be represented by an AND-tree \cite{Bru91}, 
whose structure is more intuitive. 
Roughly speaking, while an \sldp-derivation is a sequence of 
queries, in an AND-tree each node is labeled with a literal; 
when it is resolved with a (possibly probabilistic) clause 
with body literals $b_1,\ldots,b_n$, 
we add $n$ nodes as children labeled with $b_1,\ldots,b_n$
(no children if $n=0$, i.e., the clause is a fact).
W.l.o.g., we assume in the following that the initial query is 
atomic.\!\footnote{Nevertheless, one could consider 
an arbitrary query $Q$ by adding a clause of the 
form $main(X_1,\ldots,X_n) \leftarrow Q$ for some fresh 
predicate $main/n$, where $\var(Q) = \{X_1,\ldots,X_n\}$, 
and then consider query $main(X_1,\ldots,X_n)$ instead.} 

In order to represent the AND-trees of the proofs of a query
(i.e., its successful \sldp-derivations),  
we follow these guidelines:
\begin{itemize}
\item First, we backpropagate the computed mgu's 
to all queries in the considered derivation, so that
all of them become ground (a consequence of the program
being range-restricted \cite{Mug00}).
Formally, if $D$ has the form 
$\tuple{Q_0,\ch_0} \leadsto_{\sigma_1} \ldots
\leadsto_{\sigma_n} \tuple{Q_n,\ch_n}$, we consider
$\tuple{Q_0\sigma,\ch_0} \leadsto_{\sigma'_1} \ldots
\leadsto_{\sigma'_n} \tuple{Q_n\sigma,\ch_n}$
instead, where $\sigma = \sigma_1\sigma_1\ldots\sigma_n$.

\item As for the choice expressions labeling the edges
of the original derivation, we only show those associated
to the resolution of negative literals. 
The case of (probabilistic) positive atoms is considered redundant 
since the information given by an atomic choice is somehow
implicit in the tree.

\item Negative literals $\neg a$ 
have only one child, $\Box$, and the edge is 
labeled with the same choice expression as in
the \sldp-derivation, since this information cannot be
extracted from the AND-tree. 
To be more precise, in order to improve its readability, 
we choose a more intuitive representation 
for atomic choices: given an atomic choice
$\alpha = (c,\theta,i)$ with
$c = (h_1\cons p_1;\ldots;h_n\cons p_n\leftarrow B)$,
we let $\q(\alpha) = h_i\theta$. It is extended to
choice expressions in the obvious way:
$\q(\neg \ch) = \neg \q(\ch)$, 
$\q(\ch_1\wedge \ch_2) = \q(\ch_1)\wedge\q(\ch_2)$,
and $\q(\ch_1\vee\ch_2) = \q(\ch_1)\vee\q(\ch_2)$.
For instance, given the choice expression\\[1ex]
\mbox{}\hspace{0ex}
$\ch =\blue{(\neg(c_3,\sigma,1)\wedge \neg (c_4,\sigma,1)) \vee
(\neg(c_3,\sigma,1)\wedge(c_5,\sigma,1)\wedge\neg 
(c_6,\sigma,1))
}$\\[1ex]
we have $\q(\ch) =$\\[1ex]
\mbox{}\hspace{0ex}
$\blue{(\neg \ffp(p1)\wedge \neg vaccinated(p1)) \vee
(\neg \ffp(p1)\wedge vulnerable(p1)\wedge\neg young(p1))
}$\\[1ex]
Let $\q(\ch)$ be the expression labeling the edge
from $\neg a$. We further remove the occurrences of 
$\neg a$ in $\q(\ch)$ (if any) since they are clearly redundant.
\end{itemize}
\begin{figure}[t]
  \mbox{}\hspace{-3ex}
  \begin{minipage}{.37\linewidth}
  \centering
  $
  \xymatrix@C=-0.25em@R=3.5em{
	& covid(p1) \ar[dl] \ar[d] \ar[dr] \\
	contact(p1,p2) & covid(p2) \ar[d] & \neg pro(p1)
    \ar[d]^{\red{\q(\ch)}}\\
	&  pcr(p2) & \Box \\
  }
  $\\[1ex]
  \flushleft \hspace{22ex}(a)
  \end{minipage}
  \begin{minipage}{.60\linewidth}
  \centering
  $
  \begin{array}{l@{~~~~}l}
  covid(p1) & \blue{\mbox{$p1$ has covid-19 because}}\\
  \hspace{3ex}contact(p1,p2)
  &\hspace{3ex}\blue{\mbox{$p1$ had contact with $p2$}}\\
  \hspace{3ex}covid(p2) 
  &\hspace{3ex}\blue{\mbox{and $p2$ has covid-19 because}}\\
  \hspace{6ex}pcr(p2) 
  &\hspace{6ex}\blue{\mbox{the pcr test of $p2$ was positive}}\\  
  \hspace{3ex}\neg pro(p1) 
  &\hspace{3ex}\blue{\mbox{and $p1$ was not protected because}}\\ 
  \hspace{6ex}\neg \ffp(p1)
  &\hspace{6ex}\blue{\mbox{$p1$ didn't wear an ffp2 mask}}\\
  \hspace{6ex}\neg vacc(p1)
  &\hspace{6ex}\blue{\mbox{and $p1$ was not vaccinated}}\\   
  \hspace{6ex};
  &\hspace{6ex}\blue{\mbox{or because}}\\
  \hspace{6ex}\neg \ffp(p1)
  &\hspace{6ex}\blue{\mbox{$p1$ didn't wear an ffp2 mask}}\\
  \hspace{6ex}vuln(p1)
  &\hspace{6ex}\blue{\mbox{and $p1$ is vulnerable}}\\   
  \hspace{6ex}\neg young(p1)
  &\hspace{6ex}\blue{\mbox{and $p1$ is not young}}\\  
  \end{array}
  $\\[2ex]
  (b)\hspace{25ex}(c)
  \end{minipage}
\caption{Representing explanations: graphical vs textual vs natural language, where $\q(\ch) = (\neg \ffp(p1)\wedge \neg vaccinated(p1)) \vee
(\neg \ffp(p1)\wedge vulnerable(p1)\wedge\neg young(p1))$} \label{fig:representations}
\end{figure}
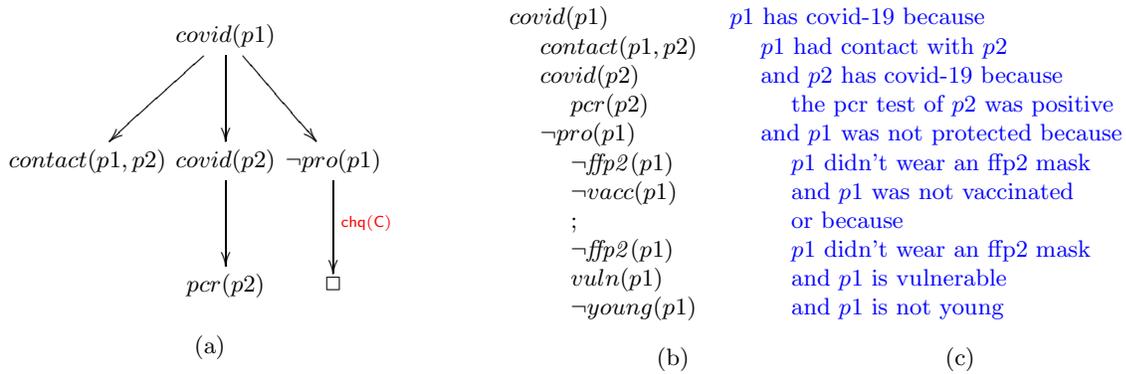
Consider, for instance, the second proof of query
$covid(p1)$ shown in Figure~\ref{fig:ex-running-neg}
that computes the choice expression $\ch_2$ 
above (\ref{eqn:c2}). The corresponding AND-tree is 
shown in Figure~\ref{fig:representations} (a).
Note that only the choice expression for
$\neg protected(p1)$ labels an edge.
An AND-tree can also be represented in textual form,
as shown in Figure~\ref{fig:representations}~(b).
Alternatively, one can easily rewrite the textual representation
using natural language. For this purpose, the user should 
provide appropriate program annotations. For instance, 
given the following annotation:\\[1ex]
\mbox{}\hspace{4ex}
\blue{\texttt{\%!read covid(A) as: "A has covid-19"}}
\\[1ex]
we could replace $covid(p1)$ by the sentence
``p1 has covid-19''. Given appropriate annotations,
the AND-tree in
Figure~\ref{fig:representations} (b) could be represented
as shown in Figure~\ref{fig:representations} (c). 
Furthermore, one can easily design an appropriate interface where
initially all elements are \emph{folded} and one should
click on each fact in order to unfold the list of reasons.
In this way, the user could more easily navigate through the
explanation and focus on her particular interests.
Another improvement could consist in showing the concrete alternatives
to negative literals in $\q(\ch)$, e.g., show
$\{surgical(p1),cloth(p1),none\}$ when clicking on
$\neg \ffp(p1)$.

A prototype implementation has been undertaken 
with promising results.

\section{Related Work} \label{sec:relwork}

We found very few works in which a query-driven inference 
mechanism for some form of 
probabilistic logic programs with negation
is formalized. Among them, the closest are the works of Riguzzi
\cite{Rig07,Rig10,Rig09}, although the aim is different to
ours (efficiently 
computing the marginal probability of a query rather
than producing comprehensible explanations).
Specifically, \cite{Rig07} proposes an algorithm for performing 
inference with LPADs where a modification of 
SLDNF-resolution is used for computing explanations 
in combination with BDDs. 
%
Later on, \cite{Rig10} presents an algorithm 
for performing inference on non-modularly acyclic LPADs. 
For this purpose, SLGAD (SLG for Annotated Disjunctions)
is introduced, an extension of SLG-resolution for LPAD.
Here, the inference mechanism 
uses tabling to avoid redundant 
computations and to avoid infinite loops. 
A distinctive feature of this approach is that the 
SLGAD-tree computes a set of composite choices which are
\emph{mutually incompatible}. This is achieved by performing
a sort of \emph{linearization} in the computation of
atomic choices, so that every time a choice is done,
a new branch where this choice is not selected is also added.
This is appropriate for 
computing the marginal probability of a query but
makes the (typically huge) trees much less useful from 
the point of view of explainability.


The closest work is \cite{Rig09}, 
which presents an extension of SLDNF-resolution for 
ICL (\emph{Independent Choice Logic} \cite{Poo97}). 
There are, however, significant differences.
First, the considered language is different
(ICL vs LPAD). Second, \cite{Rig09} aims at defining a 
technique to compute the marginal probability of a query
while our work is concerned with the generation of
comprehensible explanations. 
Finally, the shape of the resolution
trees are different since we deal with sets of composite
choices (represented by choice expressions) so that 
queries where a negated literal is selected 
have (at most) one child. Indeed, the introduction of
an algebra of choice expressions, together with 
negated atomic choices for a more compact representation,
are significant differences w.r.t.\ \cite{Rig09},
and they are essential for producing appropriate explanations.
We also note that \cite{Rig09} does not require 
grounding the program, although in return it imposes 
some very strong conditions in order to guarantee that
every time a literal is selected, it is ground and, 
morever, the computed mgu completely 
grounds the considered clause in the resolution step.

A different approach to computing explanations is 
introduced in \cite{Vid22}. 
The aim of this work is similar to ours, but there are 
significant differences too. On the one hand, the language 
considered is ProbLog without negation nor annotated 
disjunctions, so it is a much simpler setting
(it can be seen as a particular case of the language 
considered in this work). On the other hand, the 
generated explanations are \emph{programs} (a set of ground
probabilistic clauses), which are obtained through 
different unfolding transformations. In fact,
\cite{Vid22} can be seen as a complementary 
approach to the one presented here.

Finally, let us mention several approaches to improve the
generation of explanations in some 
closely related but non-probabilistic 
fields: logic programming and \emph{answer
set programming} (ASP) \cite{BET11}. 
First, \cite{CFM20} presents a tool, $\tt xclingo$, 
for generating explanations from annotated
ASP programs. Annotations are then used
to construct derivation trees containing textual explanations. 
Moreover, the language allows the user to select 
\emph{which} atoms or rules should be included in the explanations.
On the other hand, \cite{ACCG20} presents 
so-called \emph{justifications} for ASP programs with constraints,
now based on a goal-directed semantics. As in the previous work,
the user can decide the level of detail required in a
justification tree, as well as add annotations to produce
justifications using natural language.
Some of the ideas presented in Section~\ref{sec:explaining} follow
an approach which is similar to that of \cite{ACCG20}.
Other related approaches are the \emph{off-line and on-line 
justifications} of \cite{PSE09}, which provide a 
graph-based explanation of the truth value of a literal,
and the \emph{causal graph justifications}
of \cite{CFF14}, which explains why a literal is contained 
in an answer set (though negative literals are not represented).
Obviously, our work shares the aim of these papers 
regarding the generation of comprehensible explanations
in a logic setting. However, the
considered language and the applied techniques are
different. Nevertheless, we believe that our approach could 
be enriched with some of the ideas in these works.

\section{Concluding Remarks and Future Work} \label{sec:conc}

In this work, we have presented a new approach 
for query-driven inference in a 
probabilistic logic language, thus defining an 
extension of the SLDNF-resolution principle
called \sldp-resolution. 
Here, each proof of a query is now accompanied by a 
so-called \emph{choice expression} that succinctly
represents the possible worlds 
where this proof holds. 
We have also shown that choice expressions form 
a Boolean algebra, which allows us to manipulate them in a 
very flexible way. 
%
%
Furthermore, 
the generated proofs are especially appropriate to
produce comprehensible explanations for a given query. These
explanations can also be expressed using natural language
and exhibit a causal structure.

As future work, we consider several extensions. 
On the one hand, we plan to
deal with a broader class of programs. For this purpose,
we will explore the definition of an extension of
SLG-resolution \cite{CW96}
and/or some of the approaches for
goal-directed execution of ASP programs 
(e.g., \cite{MBMG12}).
On the other hand, we would also like to extend 
the inference mechanism in order to include \emph{evidences} 
(that is, ground facts whose true/false value is known).


\bibliographystyle{splncs04}

\clearpage
\appendix

\section{Technical Proofs}

First, we will prove that 
$\tuple{\widetilde{\bigch},\wedge,\vee,\neg,\top,\bot}$
is indeed a Boolean algebra. Let us recall that
$\ch_1\sim\ch_2$ if 
$\omega_{\gamma(\ch_1)} = \omega_{\gamma(\ch_2)}$
and $\widetilde{\bigch}$ denotes the quotient set of
$\bigch$ (the domain of choice expressions for a given program)
by ``$\sim$'', where
\begin{itemize}
  \item $\gamma(\bot) = \{\}$, i.e., $\bot$ denotes an
  inconsistent set of atomic choices.
  \item $\gamma(\top) = \{\{\}\}$, i.e., $\top$ represents
  a composite choice, $\{\}$, that can be extended in order
  to produce all possible selections. 
  \item $\gamma(\alpha) = \{\{\alpha\}\}$.
  \item $\gamma(\neg \ch) = \duals(\gamma(\ch))$, i.e., $\neg \ch$
  represents the complement of $\ch$.
  \item $\gamma(\ch_1\wedge\ch_2) = \mins(\gamma(\ch_1) \otimes
  \gamma(\ch_2))$.
  \item $\gamma(\ch_1\vee\ch_2) =
  \mins(\gamma(\ch_1)\cup\gamma(\ch_2))$.
\end{itemize}
Here, $\ch\in\widetilde{\bigch}$
denote the equivalence class $[\ch]$ 
when no confusion can arise.

Moreover, we recall the following lemma from \cite{Poo00}:

\begin{lemma}[Lemma 4.8 in \cite{Poo00}] \label{lemma:complement}
  Let $K$ be a set of composite choices. Then, $\duals(K)$
  is a complement of $K$.
\end{lemma}
Now, we can prove that 
$\tuple{\widetilde{\bigch},\wedge,\vee,\neg,\top,\bot}$
is a Boolean algebra.

\begin{proposition} \label{prop:bool-alg}
$\tuple{\widetilde{\bigch},\wedge,\vee,\neg,\top,\bot}$ 
is a Boolean algebra.
\end{proposition}

\begin{proof} 
We prove the axioms of a Boolean algebra. 
Here, to prove that two choice expressions, 
$\ch_1$ and $\ch_2$, are equivalent, we will 
prove that $\gamma(\ch_1)\approx\gamma(\ch_2)$, 
where $K_1 \approx K_2$ 
if $K_1 = K_2$ or $\omega_{K_1} = \omega_{K_2}$.
In the
following, we assume $\ch,\ch_1,\ch_2,\ch_3,\ldots$ are 
choice expressions 
that denote the corresponding class. Moreover,
we ignore the applications of function $\mins$ for clarity
since it cannot affect the result, only
to the element of the class which is shown.

(Associativity) The first axiom 
$\ch_1\vee (\ch_2\vee \ch_3) = (\ch_1\vee \ch_2) \vee \ch_3$ 
follows from the associativity of set union since
$\gamma(\ch_1\vee\ch_2) = \gamma(C_1)\cup\gamma(C_2)$.
Similarly, the second axiom 
$\ch_1\wedge(\ch_2\wedge \ch_3) = (\ch_1\wedge \ch_2) \wedge \ch_3$ 
follows from the fact that ``$\otimes$'' is 
associative over sets since 
$\gamma(\ch_1\wedge\ch_2) = \gamma(C_1)\otimes\gamma(C_2)$.

(Commutativity) Again, both axioms 
$\ch_1\vee \ch_2= \ch_2\vee \ch_1$ and 
$\ch_1\wedge \ch_2 = \ch_2 \wedge \ch_1$ follow 
straightforwardly from the commutativity of 
``$\cup$'' and ``$\otimes$''.

(Absorption) Axiom $\ch_1 \vee (\ch_1\wedge \ch_2) = \ch_1$ 
follows from the fact that the composite choices in 
$\gamma(\ch_1\wedge \ch_2)$ are all
redundant (i.e., do not affect to the considered class)
w.r.t.\ $\gamma(\ch_1)$ 
since they will be supersets of the
sets in $\gamma(\ch_1)$.
As for axiom $\ch_1 \wedge (\ch_1 \vee \ch_2) = \ch_1$, 
it follows from the following equivalences: 
\[
\begin{array}{l@{~~~}l}
\gamma(\ch_1 \wedge (\ch_1 \vee \ch_2))\\
= \gamma(\ch_1)\otimes(\gamma(\ch_1)\cup\gamma(\ch_2))\\
= (\gamma(\ch_1)\otimes\gamma(\ch_1))\cup(\gamma(\ch_1)\otimes\gamma(\ch_2))\\
\approx \gamma(\ch_1)\cup(\gamma(\ch_1)\otimes\gamma(\ch_2))\\
= \gamma(\ch_1)
\end{array}
\]
since all composite choices in 
$\gamma(\ch_1)\otimes\gamma(\ch_2)$ are supersets of
those in $\gamma(\ch_1)$ and, thus, do not affect 
to the considered class.

(Identity) Both axioms $\ch\vee \bot = \ch$ 
and $\ch\wedge \top = \ch$ 
are trivial by definition (e.g., 
by taking $\gamma(\bot)=\{\}$ and $\gamma(\top) = \{\{\}\}$).

(Distributivity) The first axiom, $\ch_1\vee (\ch_2 \wedge \ch_3) =
(\ch_1\vee \ch_2) \wedge (\ch_1\vee \ch_3)$ 
follows from the following equivalences: 
\[
\begin{array}{l@{~~~}l}
\gamma((\ch_1\vee \ch_2) \wedge (\ch_1\vee \ch_3))\\
= (\gamma(\ch_1)\cup\gamma(\ch_2))
\otimes
(\gamma(\ch_1)\cup\gamma(\ch_3))\\
=(\gamma(\ch_1)\otimes\gamma(\ch_1))
\cup
 (\gamma(\ch_1)\otimes\gamma(\ch_3))
 \cup
 (\gamma(\ch_2)\otimes\gamma(\ch_1))
\cup
 (\gamma(\ch_2)\otimes\gamma(\ch_3))\\
 \approx \gamma(\ch_1)\cup
 (\gamma(\ch_2)\otimes\gamma(\ch_3))\\
= \gamma(\ch_1\vee(\ch_2\wedge\ch_3))
\end{array}
\] 
The second axiom, 
$\ch_1 \wedge (\ch_2\vee \ch_3) = (\ch_1\wedge \ch_2)
\vee (\ch_1\wedge \ch_3)$
follows easily since 
\[
\begin{array}{l}
\gamma(\ch_1 \wedge (\ch_2\vee \ch_3)) \\
= \gamma(\ch_1)\otimes(\gamma(\ch_2)\cup\gamma(\ch_3)) \\
= (\gamma(\ch_1)\otimes\gamma(\ch_2))\cup(\gamma(\ch_1)\otimes\gamma(\ch_3)) \\
= \gamma((\ch_1\wedge \ch_2)
\vee (\ch_1\wedge \ch_3))
\end{array}
\]

(Complements) Axiom $\ch\vee \neg \ch = \top$ follows from
Lemma~\ref{lemma:complement}. 
Thus, $\gamma(\ch)\cup\gamma(\neg \ch)
= \gamma(\ch)\cup\duals(\gamma(\ch))$ cover all possible
selections.
Axiom $\ch\wedge \neg \ch = \bot$
holds since every element in 
$\gamma(\ch)\otimes\duals(\gamma(\ch))$ is
inconsistent by construction.
\qed
\end{proof}
The following 
auxiliary lemma is required for the remaining results:

\begin{lemma}\label{lemma:mins-hits}
  Let $K_1,K_2$ be sets of composite choices. Then,
  $\mins(\hits(K_1\otimes K_2)) = \mins(\hits(K_1)\cup\hits(K_2))$.
\end{lemma}

\begin{proof}
  Trivially, we have $\hits(K_1)\subseteq \hits(K_1\otimes K_2)$
  and $\hits(K_2)\subseteq \hits(K_1\otimes K_2)$. Then, the 
  claim follows from the fact that the hitting sets that combine
  elements from $K_1$ and $K_2$ are redundant and are 
  removed by $\mins$ since they include the hitting sets 
  in either $\hits(K_1)$ or $\hits(K_2)$. \qed
\end{proof}
Double negation elimination 
and the usual De Morgan's laws also hold for choice expressions:

\begin{proposition} \label{prop:demorgan}
Let $\ch,\ch_1,\ch_2\in \widetilde{\bigch}$. Then,
\begin{enumerate}
\item $\neg \neg \ch = \ch$;
\item $\neg (\ch_1 \vee \ch_2) = \neg \ch_1 \wedge \neg \ch_2$;
\item $\neg (\ch_1 \wedge \ch_2) = \neg \ch_1 \vee \neg \ch_2$.
\end{enumerate}
\end{proposition}

\begin{proof}
  We follow the same considerations as in the proof of
  Proposition~\ref{prop:bool-alg}.
  The proof of the double negation elimination follows
  straightforwardly from Lemma~\ref{lemma:complement} and
  the fact that the complement of a complement returns
  the original set or another one which belongs to the same
  equivalence class, i.e., 
  $\gamma(\neg\neg\ch) = \duals(\gamma(\neg\ch))
  = \duals(\duals(\gamma(\ch))) = \gamma(\ch)$.
  
  The first De Morgan's law can be proved as follows:
  \[
  \begin{array}{l}
  \gamma(\neg (\ch_1\vee \ch_2)) \\
  = \duals(\gamma(\ch_1) \cup \gamma(\ch_2))\\
  = \hits(\ol{\gamma(\ch_1) \cup \gamma(\ch_2)}) \\
  = \hits(\ol{\gamma(\ch_1)} \cup \ol{\gamma(\ch_2)})\\
  \approx \hits(\ol{\gamma(\ch_1)} \otimes \ol{\gamma(\ch_2)})
  ~~~~\mbox{(by Lemma~\ref{lemma:mins-hits})}\\
  = \hits(\ol{\gamma(\ch_1)}) \otimes \hits(\ol{\gamma(\ch_2)})\\
  = \duals(\gamma(\ch_1))\otimes\duals(\gamma(\ch_2))\\
  = \gamma(\neg \ch_1)\otimes\gamma(\neg \ch_2)\\
  = \gamma(\neg \ch_1 \wedge \neg \ch_2)
  \end{array}
  \]
  The proof of the second De Morgan's law proceeds analogously:
  \[
  \begin{array}{l}
  \gamma(\neg(\ch_1\wedge \ch_2)) \\
  = \duals(\gamma(\ch_1\wedge \ch_2))\\
  = \duals(\gamma(\ch_1) \otimes \gamma(\ch_2))\\
  = \hits(\ol{\gamma(\ch_1) \otimes \gamma(\ch_2)}) \\
  = \hits(\ol{\gamma(\ch_1)} \otimes \ol{\gamma(\ch_2)}) \\
  \approx \hits(\ol{\gamma(\ch_1)}) \cup \hits(\ol{\gamma(\ch_2)}) ~~~~\mbox{(by Lemma~\ref{lemma:mins-hits})}\\
  = \duals(\gamma(\ch_1)) \cup \duals(\gamma(\ch_2))\\
  = \gamma(\neg\ch_1)) \cup \gamma(\neg\ch_2))\\
  = \gamma(\neg \ch_1 \vee \neg \ch_2)
  \end{array}
  \]
  \qed
\end{proof}

Finally, we can prove the soundness and completeness
of \sldp-resolution, i.e., that $\expl(Q)$ indeed produces
a set of covering explanations for $Q$.

\mbox{}\\
\textbf{Theorem~\ref{th:sound}.}
\emph{
Let $\cP$ be a sound program and $Q$ a ground query. 
  Then, $\omega_s \models Q$ iff there exists a composite
  choice $\kappa\in\expl_\cP(Q)$ such that $\kappa\subseteq s$.
  }

\begin{proof}
  The proof follows a similar scheme as that of the proof of 
  Theorem~4.7 in \cite{Rig09} given the fact that
  $\neg \ch$ represents a complement of $\ch$ by definition
  and that function $\dnf$ preserves the worlds represented
  by a choice expression (Propositions~\ref{prop:bool-alg}
  and \ref{prop:demorgan}).
  
  We prove the theorem by structural induction on the set of trees
  in $\Gamma$, the \sldp-tree for $Q$.
  
  Let us first consider the base case, where $\Gamma$ only contains
  the main tree. Therefore, no negative literal has been selected.
  Consider a successful \sldp-derivation 
  $\tuple{Q,\top} = \tuple{Q_0,\top} \leadsto_{\sigma_1}
  \ldots \leadsto_{\sigma_n} \tuple{Q_n,\ch_n}$. 
  By definition,
  we have that $\ch_n = \alpha_1\wedge\ldots\wedge\alpha_m$,
  where each $\alpha_i$ is a (positive) atomic choice, 
  $i=1,\ldots,n$. Trivially, 
  $\gamma(\ch_n) = \{\{\alpha_1,\ldots,\alpha_m\}\} = \{\kappa\}$
  and $\kappa$ is consistent by construction and the fact that
  function $\dnf$ preserves the worlds 
  (Propositions~\ref{prop:bool-alg}
  and \ref{prop:demorgan}).
  Thus, the SLDNF-derivation 
  $Q_0 \leadsto_{\sigma_1} \ldots \leadsto_{\sigma_n} Q_n$
  can be proved in every world $\omega_s$ where $\kappa \subseteq s$
  and, equivalently, $\omega_s \models Q$.
  
  The opposite direction is similar. Let $\omega_s$ be a world
  such that $\omega_s \models Q$. Then, there exists an 
  SLDNF-derivation for $Q$ in $\omega_s$. Now, we can construct
  an \sldp-derivation that mimics the steps of the SLDNF-derivation
  by applying either the first or the second case in the 
  definition of \sldp-tree. If the considered predicate is 
  not probabilistic, the equivalence
  is trivial. Otherwise, let us consider that clause
  $c\theta = (h_i \leftarrow B)\theta$ from $\omega_s$ 
  has been used in the step. Therefore, the \sldp-step will
  add the atomic choice $(c,\theta,i)$ to the current
  choice expression. Hence, the computed
  choice expression in the leaf of this derivation will 
  have the form $\alpha_1\wedge\ldots\wedge\alpha_m$
  so that $\{\alpha_1,\ldots,\alpha_m\}\subseteq s$.
  
  Let us now consider the inductive case ($\Gamma$ includes more
  than one tree). By the inductive hypothesis, we assume that
  the claim holds for every \sldp-tree of $\Gamma$ 
  which is not the main tree.
  Then, for each step,
  if the selected literal is positive, the proof proceeds as in
  the base case. Otherwise, let $\neg a$ be the selected literal.
  By the inductive hypothesis,
  we have that, $\omega_s \models a$ iff there exists a composite
  choice $\kappa\in\expl_\cP(a)$ such that $\kappa\subseteq s$.
  Let $\ch^a_1,\ldots,\ch^a_j$ be the choice expressions in the
  leaves of the \sldp-tree for $a$. 
  By definition, we have that every 
  selection that extends a
  composite choice in $\gamma(\neg (\ch^a_1\vee\ldots\vee\ch^a_j))$
  is inconsistent with any selection $s$ with $\omega_s \models a$.
  Therefore, no SLDNF-derivation for $a$ can be successful
  in the worlds of $\omega_{\gamma(\neg (\ch^a_1\vee\ldots\vee\ch^a_j))}$ and
  the claim follows.
  
  Consider now the opposite direction. Let $\omega_s$ be a world
  such that $\omega_s\models Q$ and, thus, there exists a successful
  SLDNF-derivation for $Q$ in $\omega_s$. 
  As in the base case, we can construct
  an \sldp-derivation that mimics the steps of the SLDNF-derivation
  in $\omega_s$. If the selected literal is positive, the claim follows
  by the same argument as the base case. 
  Otherwise, consider a negative literal $\neg a$ which 
  is selected in some query. By the inductive hypothesis, we have
  that $\expl(a)$ is indeed a covering set of explanations for $a$.
  Moreover, by construction, 
  the negated choice
  expression, say $\neg \ch_a$ indeed represents a complement
  of this set of explanations (i.e., $\gamma(\neg \ch_a)$ is a
  complement of $\gamma(\ch_a)$). Hence, this query has a child
  with an associated choice expression, 
  $\dnf(\ch\wedge\neg \ch_a) \neq \bot$. Therefore by mimicking
  all the steps of the successful SLDNF-derivation, we end up
  with a leaf with a choice expression $\ch'$ such that 
  $s \supseteq \kappa$ for all $\kappa\in\gamma(\ch')$.
  \qed
\end{proof}

\end{document}